\documentclass[sigconf]{acmart}

\usepackage{multirow}
\usepackage{multicol}

\AtBeginDocument{%
  \providecommand\BibTeX{{%
    \normalfont B\kern-0.5em{\scshape i\kern-0.25em b}\kern-0.8em\TeX}}}

\begin{document}

\title{Face Hallucination via Split-Attention in Split-Attention Network}

%
\author{Tao Lu}
\affiliation{%
  \institution{Hubei Key Laboratory of Intelligent Robot, Wuhan Institute of Technology, Wuhan, China
  }
  \city{}
  \country{}}
\thanks{$*$ is the corresponding author (w906522992@gmail.com)}

%
\author{Yuanzhi Wang$^*$}
\affiliation{%
  \institution{Hubei Key Laboratory of Intelligent Robot, Wuhan Institute of Technology, Wuhan, China}
  \city{}
  \country{}
}
\author{Yanduo Zhang}
\affiliation{%
	\institution{Hubei Key Laboratory of Intelligent Robot, Wuhan Institute of Technology, Wuhan, China}
	\city{}
	\country{}
}
\author{Yu Wang}
\affiliation{%
	\institution{Hubei Key Laboratory of Intelligent Robot, Wuhan Institute of Technology, Wuhan, China}
	\city{}
	\country{}
}
\author{Wei Liu}
\affiliation{%
	\institution{Hubei Key Laboratory of Intelligent Robot, Wuhan Institute of Technology, Wuhan, China}
	\city{}
	\country{}
}
\author{Zhongyuan Wang}
\affiliation{%
	\institution{School of Computer Science, Wuhan University, Wuhan, China}
	\city{}
	\country{}
}

\author{Junjun Jiang}
\affiliation{%
	\institution{School of Computer Science and Technology, Harbin Institute of Technology, Harbin, China}
	\city{}
	\country{}
}
%
%
%

%


\begin{abstract}
  Recently, convolutional neural networks (CNNs) have been widely employed to promote the face hallucination due to the ability to predict high-frequency details from a large number of samples.
  However, most of them fail to take into account the overall facial profile and fine texture details simultaneously, resulting in reduced naturalness and fidelity of the reconstructed face, and further impairing the performance of downstream tasks (e.g., face detection, facial recognition).
  To tackle this issue, we propose a novel external-internal split attention group (ESAG), which encompasses two paths responsible for facial structure information and facial texture details, respectively.
  By fusing the features from these two paths, the consistency of facial structure and the fidelity of facial details are strengthened at the same time.
  Then, we propose a split-attention in split-attention network (SISN) to reconstruct photorealistic high-resolution facial images by cascading several ESAGs.
  Experimental results on face hallucination and face recognition unveil that the proposed method not only significantly improves the clarity of hallucinated faces, but also encourages the subsequent face recognition performance substantially.
  Codes have been released at \url{https://github.com/mdswyz/SISN-Face-Hallucination}
\end{abstract}




\maketitle

\section{Introduction}
Face hallucination is a domain-specific super-resolution (SR) for enhancing quality and resolution of low-resolution (LR) facial images.
Using face hallucination for restoring latent high-resolution (HR) facial images from the observed LR ones is helpful for various downstream tasks.
Thus this task plays an important role in many real-world applications, such as face detection, verification, and analysis \cite{MTCNN,mobilefacenets,frecognition,zhouqiangfaceverification}. 

Recently, convolutional neural networks (CNNs) based face hallucination methods have achieved significant improvements over conventional face hallucination methods because of its powerful feature representation capacity.
Compared with the general image, face image is a highly structured object \cite{FSRSurvey}, which is composed of facial shape model and facial shape-free model (facial texture model) \cite{AAM,AAM2}.
From this point, the CNNs based face hallucination can be divided into two categories: texture-oriented \cite{EDGAN,MTC,DPDFN,PRDRN} and shape-oriented methods \cite{FSRNet,CAG,Superfan,Superidentity}.
Texture-oriented methods aim to restore fine-grained facial texture details through deep semantic features extracted by CNN.
For example, Zhou \emph{et al.} \cite{BCCNN} firstly introduced CNNs into face hallucination and designed bi-channel convolutional neural network to learn a mapping function from LR face images to HR face images.
Lu \emph{et al.} \cite{GLFSR} proposed a global-local fusion network to refine high-frequency information, thereby recovering fine facial texture details.

Despite the great success for texture-oriented methods, they ignore facial structure information, generating face images with blurry facial structure. 
Therefore, some researchers aim to develop shape-oriented methods that utilize extra prior information (e.g., facial parsing maps, facial landmarks) to provide facial structure information, thereby helping to restore facial shape, and further use these shapes to predict missing high-frequency information.
For example, Song \emph{et al.} \cite{LCGE} used SRCNN \cite{SRCNN} to extract facial landmarks to divide facial components from LR face image and feed the five components into different branches to reconstruct corresponding components, thereby generating HR face image.
Chen \emph{et al.} \cite{FSRNet} proposed a FSRNet, which utilized facial landmarks and parsing maps to super-resolve high-fidelity facial image.
There is a noteworthy problem that extra prior information needs extra model parameters to predict, which increase the cost and difficulty of training undoubtedly, and even make it difficult for the model to fit and predict unrealistic prior information, resulting in negative influence on the subsequent reconstruction process of the shape-oriented methods.
Therefore, the shape-oriented methods fail to maintain the fidelity of facial detail restoration and consistency of facial structure simultaneously, and even corrupt texture detail due to incorrect prior information.

To address the aforementioned problems, we propose a simple but efficient end-to-end method to reconstruct high-quality and high-fidelity HR face images.
Specifically, we first propose an internal-feature split attention (ISA) to improve internal correlation of feature by splitting and classifying the channel dimension of feature, which enables the network focuses on information-rich regions and pay less attention to less information-rich regions.
ISA can be defined as internal-feature attention because it improves internal correlation of features.
Then, we design an internal-feature split attention block (ISAB) to focus on restoration of facial texture details by using proposed ISA.
An external-internal split attention group (ESAG) is proposed to simultaneously maintain the consistency of facial structure and the fidelity of facial detail restoration by fusing multi-path features, where one path cascades several ISABs to focus on facial texture details and the other path focuses on the facial structure information through ISA. 
The process of multi-path feature fusion can be defined as external-feature attention because of improving external correlation between multi-path features.
Finally, we propose a split-attention in split-attention network (SISN) to cascade several ESAGs for reconstructing photorealistic HR face images.
The contributions of our work can be summarized as follows:

\begin{figure*}[h]
	\centering{\includegraphics[width=15.5cm]{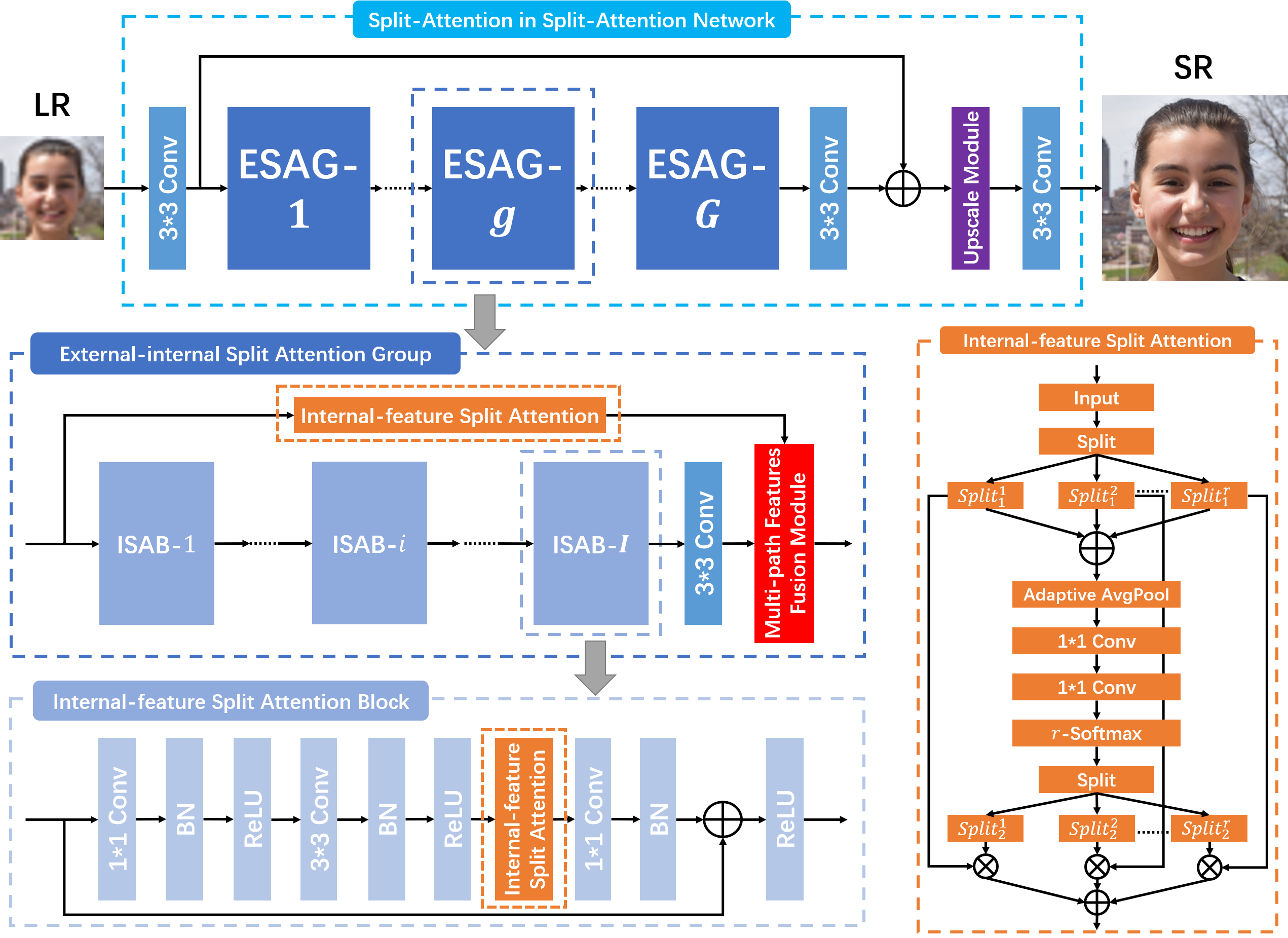}}
	\caption{Network architecture of SISN, which consists of four parts: coarse feature extraction layer, split-attention in split-attention deep feature extraction, upscale module, and reconstruction module.}
	\label{fig:SISN}
	\vspace{-0.4cm}
\end{figure*}

(\romannumeral1) We propose an ISA to enable network to focus on information-rich regions and ignores information-poor regions by splitting and classifying the channel dimension of feature, and design an ISAB to focus on restoration of facial texture details by ISA.

(\romannumeral2) To fully maintain the consistency of facial structure and the fidelity of facial detail restoration, an ESAG is proposed to fuse multi-path features responsible for facial texture details and facial structure information, respectively.

(\romannumeral3) We design a SISN to reconstruct the high-fidelity and high-quality HR face image from the input LR image by cascading several ESAGs, and demonstrate that the proposed method can effectively maintain the fidelity of facial detail restoration and consistency of facial structure through face reconstruction experiments under two public face datasets and face recognition experiments under a real-world surveillance scenarios dataset.

\section{Related work}
Face hallucination, also known as face super-resolution, was first proposed by Baker and Kanade \cite{Hallucinating}, which is the pioneer of face hallucination technique.
Since then many methods were proposed to improve the performance of hallucinating face.
Some early techniques have been proposed to super-resolve LR face image by exploring and exploiting the prior of global or local facial structures.
Therefore, they can be roughly classified into global based methods \cite{eigen,Globalorthogonal} and local based methods \cite{local1,local2,Locality}.
Global based methods super-resolve the whole image together to maintain the global structure information.
Zhou \emph{et al.} \cite{Globalorthogonal} used orthogonal canonical correlation analysis to achieve global face reconstruction.
Local based methods divide image into several overlapped small patches for better local representation ability.
Chang \emph{et al.} \cite{local1} designed locally linear representation for exploring the complex relationship between LR and HR images.
These methods fail to produce satisfactory results when faced with high upscale factor because of weak feature representation capability.

Recently, CNNs has been widely used in various upstream and downstream visual tasks  \cite{RCAN,Defogging,Defogging2,SPMF,MCN} because of its powerful feature representation capabilities. 
Naturally, CNNs based face hallucination methods \cite{EDGAN,FSRNet,LCGE,PRDRN,MTC,DPDFN,GLFSR} have achieved satisfactory performance, and have been further driven to explore higher upscale factors.
In the latest research, attention mechanism is introduced into the CNNs based methods allowing deep neural networks to pay more attention to useful information \cite{attributeattention,CAG,MTC,SAAN,Multi-ScaleAttention,SPARNet}.
For example, Xin \emph{et al.} \cite{attributeattention} designed a attribute attention network to extract and exploit latent representation of face attributes, thereby enhancing reconstruction performance for face image.
Kalarot \emph{et al.} \cite{CAG} proposed facial component-wise attention maps, which utilized extra facial prior information to allow network focusing on face feature inherent patterns.
Chen \emph{et al.} \cite{SPARNet} used spatial attention mechanism to focus on different face structures, and produced high-quality face image.

However, the above attention based methods have a serious problem. 
They used attention mechanism to either focus on the facial texture information or on the facial structure information to use the structural information to constrain texture restoration.
Therefore, these methods fail to focus on facial texture details and facial structure information at the same time, resulting in unclear and unnatural face images.

\section{Split-Attention in Split-Attention Network for Face hallucination}
\label{sec:method}

\subsection{Network Architecture}
In this subsection, we describe the network architecture of SISN in detail.
Figure. \ref{fig:SISN} shows the architecture of SISN, which mainly consists four parts: coarse feature extraction, split-attention in split-attention deep feature extraction, upscale module, and reconstruction module. Let’s denote $I_{LR}$, $I_{SR}$ as the input and output of SISN, and we use only one convolutional layer with a kernel size of 3$\times$3 to extract the coarse feature $F_{CF}$ from the input images.
\begin{equation}
F_{CF} = H_{CF}(I_{LR}),
\end{equation}
where $H_{CF}(.)$ denotes the coarse feature extracting operation with one convolutional layer.
$F_{CF}$ is then used for split-attention in split-attention deep feature extraction. So we can further have
\begin{equation}
F_{DF} = H_{DF}(F_{CF}),
\end{equation}
where $H_{DF}(.)$ denotes the deep feature extraction structure, which contains $G$ ESAGs. After the $F_{CF}$ passes through deep feature extraction and the network gets the deep feature being denoted $F_{DF}$, the upscaling operation needs to be performed by the upscale module.
\begin{equation}
F_{UP} = H_{UP}(F_{DF}),
\end{equation}
where $F_{UP}$ and $H_{UP}(.)$ denote the upscaled feature and a upscale module respectively. Finally, the upscaled feature is then reconstructed via a reconstruction module. The reconstructed $I_{SR}$ is formulated as:
\begin{equation}
I_{SR} = H_{Recon}(F_{UP}) = H_{SISN}(I_{LR}),
\end{equation}
where $H_{Recon}(.)$ and $H_{SISN}(.)$ denote the reconstruction module composed of a convolutional layer with a kernel size of 3$\times$3 and the function of our SISN, respectively.

Then SISN is optimized with loss function.
To trade off the PSNR and image construction quality, the proposed method utilizes Mean Absolute Error (MAE), namely $L_{1}$ loss function, to define our loss function. Given a training set $\left\{I_{LR}^i,I_{HR}^i\right\}_{i=1}^N$, which contains $N$ LR inputs and their HR counterparts. The $L_{1}$ loss function of SISN can be represented as:
\begin{equation}
L\left( \theta \right) = ~\frac{1}{N}{\sum\limits_{i = 1}^{N}\left\| {H_{SISN}(I_{LR}^{i}) - I_{HR}^{i}} \right\|_{1}},
\end{equation}
where $\theta$ denotes the parameter set of our network.

\subsection{Internal-feature Split Attention Block with Internal-feature Attention}
Previous attention based methods \cite{SPARNet,SAAN,MTC} used large-scale residual blocks to extract fine-grained features, which made the these methods getting better performance. 
But the limited receptive-field size and lacking of cross-channel interaction for these residual blocks also reduce the internal correlation of features. 
Inspired by previous work \cite{resnest}, we propose the ISA to improve internal correlation of features by splitting and classifying the channel dimension of feature, which enable the network focuses on information-rich regions and ignore information-poor regions. 
Then, we design an ISAB as the basic block of deep feature extraction to focus on restoration of facial texture details.  
The detailed structure of ISA and ISAB are shown in the Figure. \ref{fig:SISN}.

For the ISA, the input feature is first divided into $r$ splits along the channel axis, and the number of feature channels is defined as $C$, thus the number of channels per split is $\frac{C}{r}$.
Then these splits are fused via an element-wise summation across multiple splits.
\begin{equation}
F_{fusion} = H_{Sum}( {Split_{1}^{1}}, {Split_{1}^{2}}, \cdots ,  Split_{1}^{r}),
\end{equation}
where $H_{Sum}(.)$ denotes an element-wise summation operation, $Split_{1}^{r}$ denotes the $r^{th}$ split by the first division, and $F_{fusion}$ denotes the output by $H_{Sum}(.)$.
The $F_{fusion}$ is then passed through an adaptive average pooling layer and two 1$\times$1 convolutional layers, and is classified into $r$ splits again by a $r-softmax$ function according to the information richness of each previous split.
Each of the current $r$ splits is multiplied by each of the previous $r$ splits using an element-wise product operation respectively, thereby improving the internal correlation of features in channel-level.
Finally, we use an element-wise summation to fuse these splits as the output of ISA, which is denoted as $F_{ISA}$. $F_{ISA}$ is formulated as:
\begin{equation}
\begin{split}
F_{ISA} = H_{Sum}(  H_{MP}({Split}_{2}^{1}, {Split}_{1}^{1}), H_{MP}({Split}_{2}^{2}, \\{Split}_{1}^{2}),\cdots, H_{MP}({Split}_{2}^{r}, {Split}_{1}^{r})) = H_{ISA}( F_{input}^{ISA}),
\end{split}
\end{equation}
where $H_{MP}(.)$ denotes an element-wise product operation, the $H_{ISA}(.)$ represents the function of ISA, and $F_{input}^{ISA}$ denotes the input of ISA.

\begin{figure*}[h]
	\centering{\includegraphics[width=16cm]{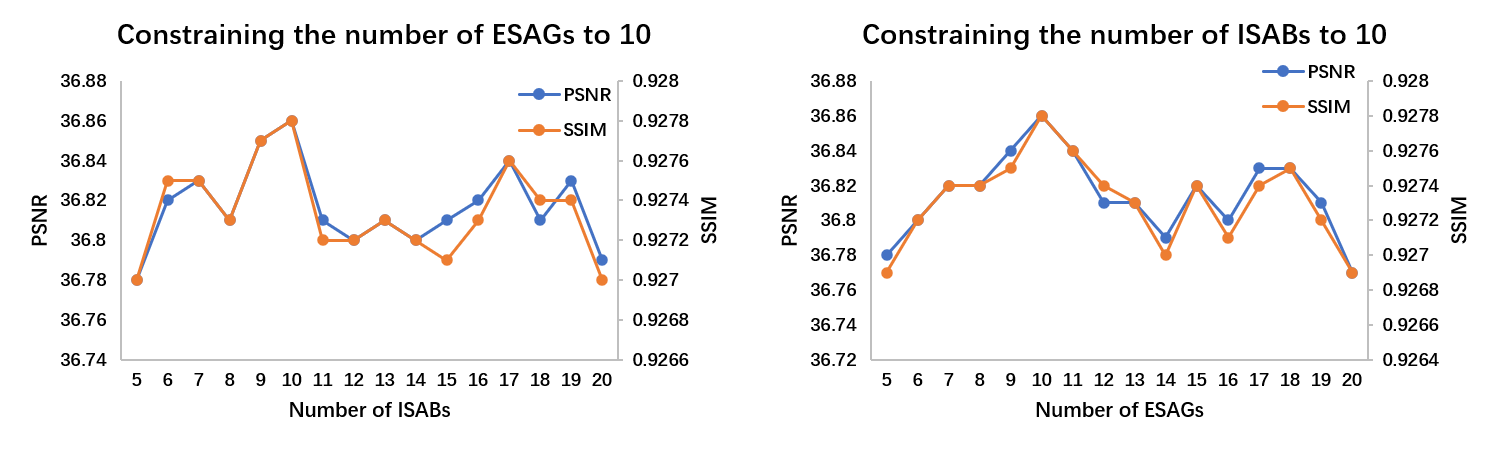}}
	\caption{Comparison with different combination of ESAGs and ISABs for 4$\times$ SR on FFHQ testing dataset. The best result is available when ESAGs is 10 and ISABs is 10.}
	\label{fig:Ablation}
\end{figure*}

\subsection{External-internal Split Attention Group with Internal- and External-feature Attention}
We now describe the proposed ESAG, which contains two paths.
The first path consists of $I$ ISABs and a convolution layer, which is mainly focus on restoring fine facial details because it produces rich semantic information. The output $F_{path}^{1}$ from this path is formulated as:
\begin{equation}
F_{path}^{1} = H_{Conv}(H_{ISAB}^{I}(H_{ISAB}^{I- 1}( \cdots H_{ISAB}^{1}( F_{input}^{ESAG} ) \cdots ))),
\end{equation}
where $H_{Conv}(.)$ and $H_{ISAB}^{I}(.)$ denote a convolutional layer with kernel size of 3$\times$3 and $I^{th}$ ISAB, respectively. $F_{input}^{ESAG}$ denotes the input of ESAG.

Considering the shallower features can provide more precise facial structure information, for an ESAG, the input feature $F_{input}^{ESAG}$ can be used as a shallow feature. 
Therefore, the second path uses $F_{input}^{ESAG}$ as input and utilizes ISA to focus on facial structure information.
The output $F_{path}^{2}$ from the second path is formulated as:
\begin{equation}
F_{path}^{2} = H_{ISA}(F_{input}^{ESAG}).
\end{equation}
Finally, we use a multi-path features fusion module to fuse the features from two paths to improve the external correlation of features, thereby restoring fine facial details while maintaining the consistency of the facial structure. The fused feature $F_{ESAG}$ is represented as:
\begin{equation}
F_{ESAG} = H_{MFM}(F_{path}^{1},F_{path}^{2}) = H_{ESAG}(F_{input}^{ESAG}),
\end{equation}
where $H_{MFM}(.)$ denotes a multi-path features fusion module using element-wise summation operation, and $H_{ESAG}(.)$ represents the function of ESAG.

\begin{figure*}[h]
	\centering{\includegraphics[width=16cm]{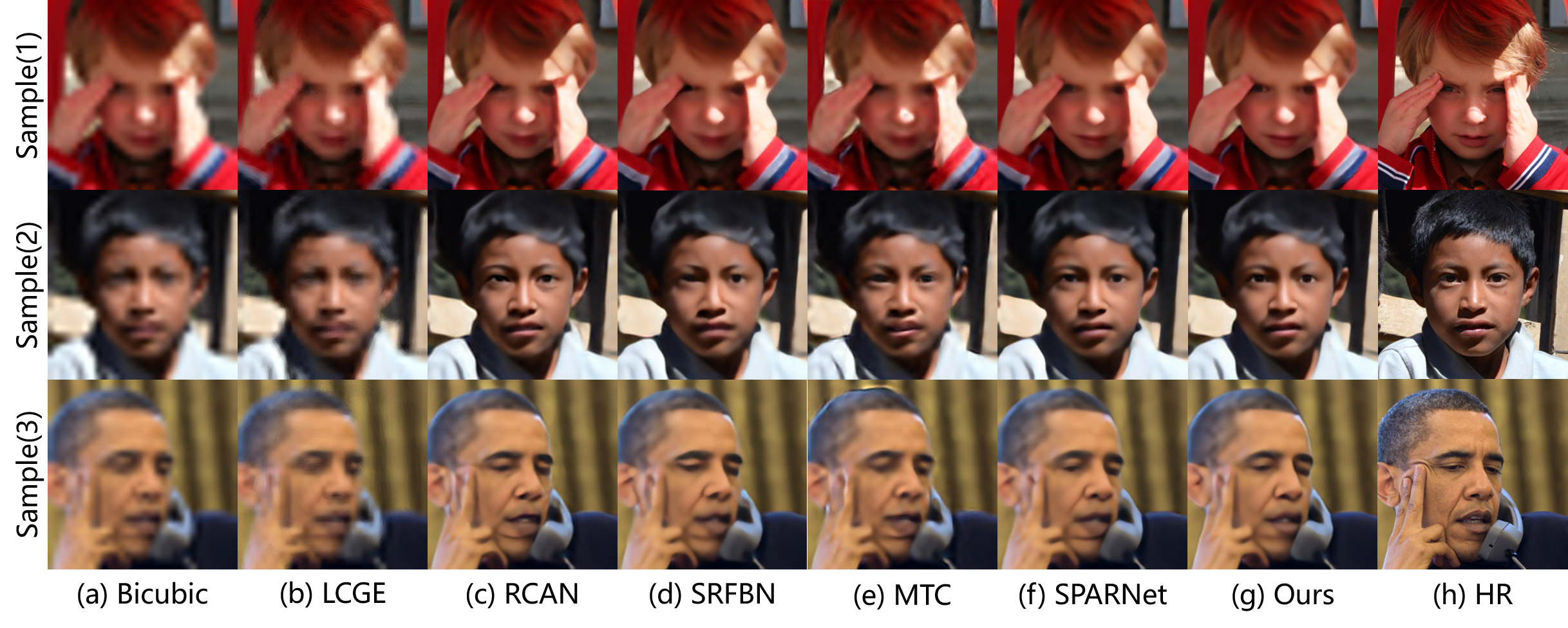}}
	\caption{Visual comparison for 8$\times$ SR with different methods. Three images from the FFHQ testing set (HR size is 256$\times$256 pixels) are selected as the test samples to show the reconstruction results of face images.}
	\label{fig:test sample_FFHQ}
	\vspace{-0.4cm}
\end{figure*}

\section{EXPERIMENTS}
\label{sec:experiments}
\subsection{Dataset and Implementation Details}
In this paper, two publicly available face datasets, namely FFHQ \cite{FFHQ} and CelebA \cite{CelebA}, are used for extensive reconstruction experiments.
For FFHQ dataset, as a novel face dataset, it consists of 70,000 high-quality images with considerable variation in terms of facial attributes such as age and ethnicity as well as image background.
We randomly select 850 images as the training dataset, 100 images as the validation dataset and 50 images as the testing dataset. 
The HR image size is 1024$\times$1024 pixels, and the downsampling factor are the 4 and 8, so that the LR image size are 256$\times$256 and 128$\times$128 pixels respectively. 
For CelebA dataset, as a large-scale face attributes dataset, it is consist of more than 200k face images.
We randomly select 2,000 images as the training dataset, 140 images as the validation dataset and 60 images as the testing dataset. 
All HR images are resized to 216$\times$176 pixels.

Data augmentation is performed on the all training images, which are randomly rotated by 90\textdegree, 180\textdegree, 270\textdegree, and flipped horizontally. 
All experiments are trained by Adam optimizer with $\beta_{1} = 0.9$, $\beta_{2} = 0.999$, and $\epsilon = 10^{-8}$.  The initial leaning rate is set to $10^{-4}$, and then decreases to half every 50 epochs.  We use PyTorch to implement our models with a RTX 2070 Super GPU.

\subsection{Evaluation Metrics}
The SR results are evaluated with four evaluation metrics: Peak Signal to Noise Ratio (PSNR), Structural Similarity (SSIM) \cite{SSIM}, Learned Perceptual Image Patch Similarity (LPIPS) \cite{LPIPS}, and Mean Perceptual Score (MPS) \cite{AIM2020}.
PSNR and SSIM are standard evaluation metrics, which are widely used in low-level visual tasks.
LPIPS and MPS are the novel perceptual metrics to measure the perceptive quality of image, where the smaller value of LPIPS means more perceptual similarity.
MPS is the average of the SSIM and LPIPS, which is formulated as:
\begin{equation}
MPS = 0.5\times(SSIM+(1-LPIPS)).
\end{equation}

\begin{table}[t]
	\begin{center}
		\caption{Verifying the effect of the two paths in ESAG. "SISN w/o structure" indicates that the path of focus on facial structure is removed. "SISN w/o detail" indicates that the ISA from all ISABs are removed.}\label{tab:Verifying the effect of the two paths}
		\begin{tabular}{c|cccc}
			\hline
			Methods   &  PSNR & SSIM & LPIPS & MPS   \\
			\cline{2-5}
			\hline
			\hline
			SISN w/o structure &36.70& 0.9260 & 0.2282 &  0.8489 \\
			SISN w/o detail &36.59 & 0.9248 & 0.2296 &  0.8476  \\
			SISN & \textbf{36.86} & \textbf{0.9278} & \textbf{0.2212} & \textbf{0.8533}  \\
			\hline
		\end{tabular}
	\end{center}
\vspace{-0.4cm}
\end{table}

\subsection{Ablation Experiments}

\subsubsection{Verifying the effect of the two paths in ESAG}
To demonstrate the role of the two paths in ESAG, we set up a group of experiments to verify.
First, we remove the path that focuses on facial structure based on SISN.
Second ,we remove the ISA from all ISABs based on SISN (equivalent to removing the core part of ISABs).
The results of this group of ablation experiments for 4$\times$ SR on the FFHQ face testing dataset (HR images are 1024$\times$1024 pixels) are shown in Table \ref{tab:Verifying the effect of the two paths} , where we can conclude that when one of the paths is removed, there is a considerable decrease in performance, which proves that the two paths have a key role in the improvement of the reconstruction performance.
\subsubsection{Exploring the best combination of ESAGs and ISABs}
In this section, we conduct the ablation studies on the FFHQ face testing dataset (HR images are 1024$\times$1024 pixels) to explore the most effective combination of ESAGs and ISABs.
We design two sets of experiments under different constraints for 4$\times$ SR. 
Inspired by previous work \cite{RCAN}, in the first set of experiments, we constrain the number of ESAGs to 10, and test the reconstruction performance of SISN when the number of ISABs is from 5 to 20 and step is 1. In the second set of experiments, we constrain the number of ISABs to 10, and test the reconstruction performance of SISN when the number of ESAGs is from 5 to 20 and step is 1. The total number of the above two sets of experiments is 31.

Figure. \ref{fig:Ablation} shows the experiment results of SISN with above two sets of experiments, we can find that when ESAGs is 10 and ISABs is 10, the performance reaches the best.
Meanwhile, the fluctuation range of PSNR (dB) and SSIM in these experiment results are small, which also proves that the proposed method does not rely on the deepening of the network to obtain better performance.

\begin{table}[t]
	\begin{center}
		\caption{Comparison for 4$\times$ and 8$\times$ SR with the state-of-the-arts on FFHQ dataset, and perform experiments on 1024$\times$1024 images. The bold denotes the best results. }\label{tab:ffhq1024}
		\begin{tabular}{c|c|cccc}
			\hline
			Methods & Scales  &  PSNR & SSIM & LPIPS & MPS   \\
			\cline{3-6}
			\hline
			\hline
			Bicubic & \multirow{7}{*}{$\times4$} & 34.80 & 0.9037 & 0.3104 & 0.7967   \\
			LCGE \cite{LCGE} && 35.26 & 0.9042 & 0.2370 & 0.8336  \\
			RCAN \cite{RCAN} && 36.59 & 0.9250 & 0.2309 & 0.8471  \\
			SRFBN \cite{SRFBN}& &36.53
			& 0.9243 & 0.2298 & 0.8473  \\
			MTC \cite{MTC}& & 36.62 & 0.9259 & 0.2249 & 0.8505  \\
			SPARNet \cite{SPARNet}& &36.37&0.9240& 0.2323 &0.8459 \\
			SISN(Ours) && \textbf{36.86} & \textbf{0.9278} & \textbf{0.2212} & \textbf{0.8533}  \\
			\hline
			\hline
			Bicubic & \multirow{7}{*}{$\times8$} & 31.34 & 0.8495 & 0.4918 & 0.6789 \\
			LCGE \cite{LCGE} && 31.18 & 0.8403 & 0.4040 & 0.7182 \\
			RCAN \cite{RCAN} && 33.15 & 0.8723 & 0.3776& 0.7474 \\
			SRFBN \cite{SRFBN}&& 32.99 & 0.8701& 0.3810 & 0.7446\\
			MTC \cite{MTC}& & 32.94 & 0.8695 & 0.3794 & 0.7451\\
			SPARNet \cite{SPARNet}&&32.57 &0.8684 & 0.3785 & 0.7450\\
			SISN(Ours) && \textbf{33.25} & \textbf{0.8738} & \textbf{0.3674} & \textbf{0.7532}\\
			\hline
		\end{tabular}
	\end{center}
\vspace{-0.4cm}
\end{table}

\begin{table}[t]
	\begin{center}
		\caption{Comparison for 4$\times$ and 8$\times$ SR with the state-of-the-arts on FFHQ dataset, and perform experiments on 256$\times$256 images. The bold denotes the best results. }\label{tab:ffhq256}
		\begin{tabular}{c|c|cccc}
			\hline
			Methods & Scales  &  PSNR & SSIM & LPIPS & MPS   \\
			\cline{3-6}
			\hline
			\hline
			Bicubic & \multirow{7}{*}{$\times4$} & 29.82 & 0.8459 & 0.3361 & 0.7549  \\
			LCGE \cite{LCGE} && 31.09 & 0.8674 & 0.2189 & 0.8243  \\
			RCAN \cite{RCAN} && 32.65 & 0.8980 & 0.1720 & 0.8630  \\
			SRFBN \cite{SRFBN}& &32.40& 0.8947 & 0.1742 & 0.8603  \\
			MTC \cite{MTC}& & 31.99 & 0.8888 & \textbf{0.1628} & 0.8630  \\
			SPARNet \cite{SPARNet}& &32.36 &0.8933 & 0.1878 & 0.8528\\
			SISN(Ours) && \textbf{32.83} & \textbf{0.9011} & 0.1664 & \textbf{0.8674}  \\
			\hline
			\hline
			Bicubic & \multirow{7}{*}{$\times8$} & 25.99 & 0.7313 & 0.5594 & 0.5860 \\
			LCGE \cite{LCGE} && 26.13 & 0.7326 & 0.4159 & 0.6584 \\
			RCAN \cite{RCAN} && 28.17 & 0.7932 & 0.3329& 0.7302 \\
			SRFBN \cite{SRFBN}& & 28.07 & 0.7906 & 0.3388 & 0.7259 \\
			MTC \cite{MTC}& & 27.57 & 0.7767 & 0.3537 & 0.7115 \\
			SPARNet \cite{SPARNet}&& 28.20 & 0.7965 & 0.3355& 0.7305\\
			SISN(Ours) && \textbf{28.37} & \textbf{0.7999} & \textbf{0.3230} & \textbf{0.7385}\\
			\hline
		\end{tabular}
	\end{center}
\vspace{-0.4cm}
\end{table}

\begin{table}[h]
	\begin{center}
		\caption{Comparison for 4$\times$ and 8$\times$ SR with the state-of-the-arts on CelebA dataset. The bold denotes the best results. }\label{tab:celeba}
		\begin{tabular}{c|c|cccc}
			\hline
			Methods & Scales  &  PSNR & SSIM & LPIPS & MPS   \\
			\cline{3-6}
			\hline
			\hline
			Bicubic & \multirow{7}{*}{$\times4$} & 28.82 & 0.8330  & 0.3589 & 0.7371  \\
			LCGE \cite{LCGE} && 27.22 & 0.6818 & 0.5379 & 0.5720  \\
			RCAN \cite{RCAN} &&32.19  & 0.8993 & 0.1876 &  0.8559 \\
			SRFBN \cite{SRFBN}& & 32.12 & 0.8981 & 0.1895 & 0.8543  \\
			MTC \cite{MTC}& & 31.92 & 0.8947 & 0.1910 & 0.8519   \\
			SPARNet \cite{SPARNet}& &31.99 & 0.8962 & 0.1965 & 0.8499 \\
			SISN(Ours) && \textbf{32.34} & \textbf{0.9009} & \textbf{0.1852} & \textbf{0.8579}  \\
			\hline
			\hline
			Bicubic & \multirow{7}{*}{$\times8$} & 24.74 & 0.6969 & 0.6290 & 0.5340 \\
			LCGE \cite{LCGE} && 23.73 & 0.5620 & 0.8356 & 0.3632 \\
			RCAN \cite{RCAN} && 27.32 & 0.7834 & 0.3842 & 0.6996 \\
			SRFBN \cite{SRFBN}& & 27.22 & 0.7810 & 0.3534 & 0.7138 \\
			MTC \cite{MTC}& & 26.91 & 0.7712 & 0.3566 & 0.7073 \\
			SPARNet \cite{SPARNet}&& 27.29 & 0.7884 & 0.3573 & 0.7156 \\
			SISN(Ours) && \textbf{27.59} & \textbf{0.7912} & \textbf{0.3496} & \textbf{0.7208}\\
			\hline
		\end{tabular}
	\end{center}
\vspace{-0.4cm}
\end{table}

\begin{figure*}[h]
	\centering{\includegraphics[width=16cm]{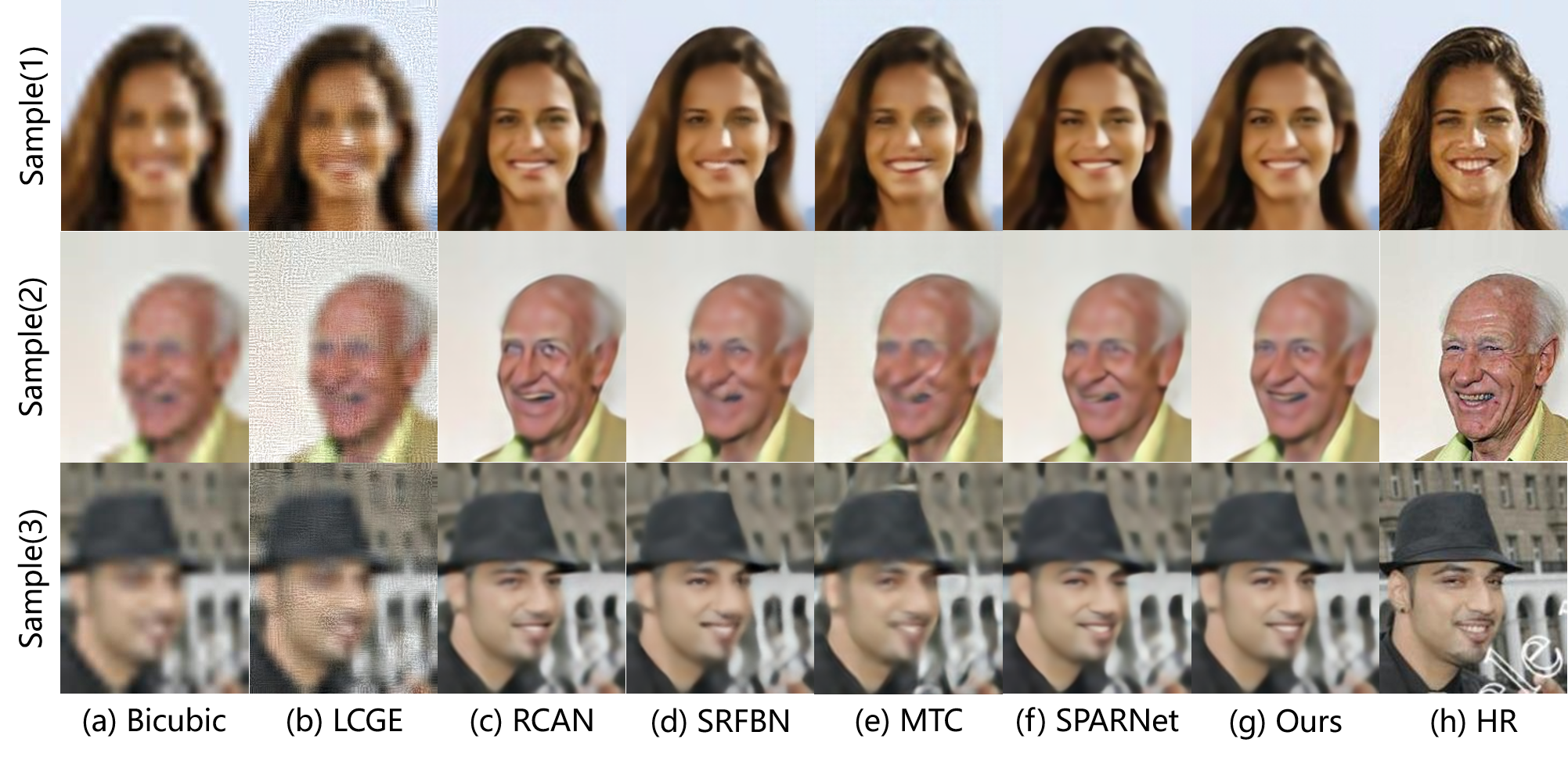}}
	\caption{Visual comparison for 8$\times$ SR with different methods. Three images from the CelebA testing set are selected as the test samples to show the reconstruction results of face images.}
	\label{fig:test sample_CelebA}
	\vspace{-0.4cm}
\end{figure*}

\subsection{Comparison with State-of-the-Arts}

In this section, we select some excellent face hallucination or general SR methods as: Bicubic, LCGE \cite{LCGE}, RCAN \cite{RCAN}, SRFBN \cite{SRFBN}, MTC \cite{MTC}, and SPARNet \cite{SPARNet}.
Bicubic is a classic image interpolation algorithm; LCGE is a face hallucination using facial landmarks; RCAN is a deep residual channel attention network based SR method; SRFBN is a feedback network based SR method; MTC is a novel face hallucination that uses the texture-attention mechanism; SPARNet is the latest face hallucination based on spatial attention mechanism.

\subsubsection{Comparing on FFHQ dataset (HR images are 1024$\times$1024 pixels)}
Table \ref{tab:ffhq1024} lists the quantitative experimental results of different state-of-the-arts and proposed method on 1024$\times$1024 FFHQ testing set for 4$\times$ and 8$\times$ SR, it is obviously that our SISN outperforms state-of-the-arts in PSNR, SSIM, LPIPS, and MPS significantly.
These results prove that SISN can improve the reconstruction performance of face images effectively by improving the consistency of facial structure and the fidelity of facial details.

For the visual performance, each algorithm has a little gap in visual reconstruction results since the size of HR face images are large. 
To further prove the effectiveness of the proposed method, we will discuss experiments on small-scale face images in the next part.

\begin{figure*}[h]
	\centering{\includegraphics[width=15cm]{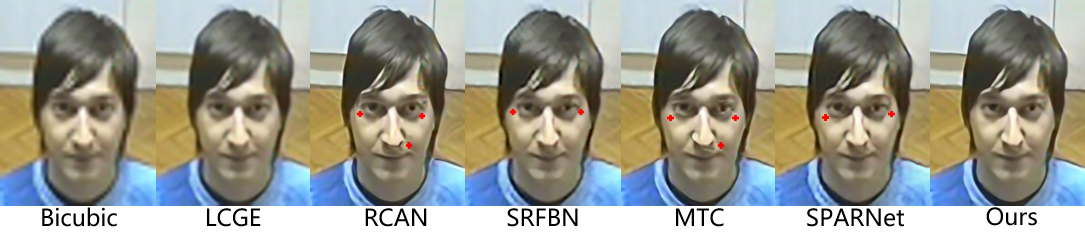}}
	\caption{Visual reconstruction results on real-world surveillance scenarios for 4$\times$ SR. Intuitively, the results of Bicubic, LCGE, and SRFBN are very vague. There are texture distortion and unrealistic artifacts around the \textcolor[rgb]{ 1,  0,  0}{red mark}. In contrast, our method produces cleaner face and maintains more facial structure information. }
	\label{fig:realworld}
	\vspace{-0.4cm}
\end{figure*}

\subsubsection{Comparing on FFHQ dataset (HR images are 256$\times$256 pixels)}
In the real-world surveillance scenarios, the resolution of the face image captured by the imaging sensor is often lower than the size of 100$\times$100, or even lower than 50$\times$50. 
At this resolution, the various downstream tasks related to face will be almost collapsed. 
Therefore, the performance of face hallucination on small-scale face images is important particularly.
To verify the performance of the proposed method on small-size face images, we resize the original HR image to 256$\times$256 as the groundtruth, the corresponding LR image size for 4$\times$ and 8$\times$ SR are 64$\times$64 and 32$\times$32 pixels, respectively.

Table \ref{tab:ffhq256} summarizes the quantitative results of different methods on the 256$\times$256 FFHQ testing set for 4$\times$ and 8$\times$ SR, it is no surprise that SISN still surpasses these state-of-the-arts, which prove that the proposed method has high generalization ability in small-scale face images.

In terms of visual performance, Figure. \ref{fig:test sample_FFHQ} shows the qualitative results of different methods for 8$\times$ SR on 256$\times$256 FFHQ testing set, we select three representative samples to compare.
(a) to (f) are the results from six selected state-of-the-art algorithms, and (h) is the HR groundtruth as the benchmark.
From these visual results, we can observe that the results of LCGE is almost same as Bicubic, and the facial texture details are almost lost.
For the sample (1), we discover intuitively that all of these state-of-the-arts cause severe structural and textural distortions in eye region, resulting in abnormal human eyes.
RCAN, SRFBN, and SPARNet lost some details in the left eye region of sample (2), while MTC causes the distortion of structural information in the eye and mouth regions.
In contrast, our method achieves the high visual performance and maintains the satisfactory facial structure.

\subsubsection{Comparing on CelebA dataset}
On the FFHQ dataset, our method shows high reconstruction performance. 
Compared with the FFHQ dataset, the original images of the CelebA dataset are blurred and contain some noise, which are closer to the real-world scenarios.
Therefore, in this section, we use this challenging dataset for experiments.
We resize the original image to 176$\times$216 as the groundtruth, the corresponding LR image size for 4$\times$ and 8$\times$ SR are only 44$\times$54 and 22$\times$27 pixels, respectively.

Table \ref{tab:celeba} shows the quantitative results of different methods on CelebA dataset, the proposed SISN has achieved an overwhelming victory in all evaluation metrics.
Figure \ref{fig:test sample_CelebA} shows the qualitative results of different methods for 8$\times$ SR on CelebA testing set.
We can conclude that compared with other methods, our SISN can still restore fine facial details while maintaining better facial structure consistency.

It is worth noting that both quantitative and qualitative experiments of LCGE have caused extremely poor results, and we analyze the potential reason that LCGE can not adapt to these images containing noise and blur, thereby resulting in the reconstructed image containing a large number of unrealistic artifacts, which can be seen in (b) of Figure \ref{fig:test sample_CelebA}.

\subsection{Face Reconstruction and Recognition on Real-world Surveillance scenarios}

\subsubsection{Face Reconstruction}
All experiments in the previous section are in the simulation scenarios, which can not simulate the real-world scenarios well.
For face hallucination, real-world surveillance scenarios are a challenging environment. Compared with the simulation scenarios, the surveillance scenarios face two serious problems: (1) there is no corresponding HR images; (2) the face image captured by the surveillance camera often contains a lot of noise and serious distortion of color and texture.
Therefore, in this scenarios, it is a severe challenge to face hallucination and requires the algorithm to reconstruct the HR face image with good facial structure and texture details from the distorted LR image as much as possible to improve the performance of downstream tasks such as face detection and recognition.

To verify the reconstruction performance of our method in real-world surveillance scenarios, we select a low-quality face image (image size is 42$\times$56 pixels) from SCface dataset \cite{SCface} which contains a large number of face images captured by surveillance cameras to perform face reconstruction experiments.
Visual result of different methods (4$\times$ SR) is shown in Figure. \ref{fig:realworld}. 
We can conclude that compared to these state-of-the-arts, our SISN produces cleaner face image and maintains more facial structure information, which can improve the performance of downstream tasks effectively and will be proved in the next part.

\begin{figure}[t]
	\centering{\includegraphics[width=5cm]{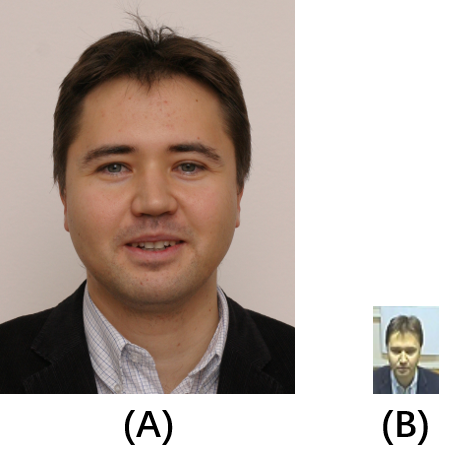}}
	\caption{A sample pair of Candidate 1, where (A) denotes a high-definition frontal face image captured by a SLR camera as a face feature sample of Candidate 1, and (B) denotes a LR face image captured from a surveillance camera as a target image of the face recognizer.} 
	\label{fig:sample_facerec}
	\vspace{-0.4cm}
\end{figure}

\begin{table*}[t]
	\begin{center}
		\caption{Comparison results for face similarity of face images reconstructed by different methods. We randomly select high-definition frontal face images of six test candidates captured by SLR camera in the SCface dataset as source face feature samples, and select the LR face images captured by the same surveillance camera corresponding to the six test candidates as the target image of the face recognizer. The bold denotes the best similarity, and the "-" means that the face of candidate is not detected.}\label{tab:facerec}
		\begin{tabular}{c|c|c|c|c|c|c|c}
			\hline
			Methods & Candidate 1  &  Candidate 2 & Candidate 3 & Candidate 4 & Candidate 5  & Candidate 6 & Average Similarity \\
			\hline
			\hline
			LR image & - & - & - & - & - & 0.449837 & 0.074973 \\
			LCGE \cite{LCGE} & 0.468556 &0.222741  & 0.269097 & 0.106820 & - & 0.607645 &0.279143 \\
			RCAN \cite{RCAN} &0.356093 & 0.590624 & 0.509804 & 0.512736 &  0.347659 & 0.612440 & 0.488226\\
			SRFBN \cite{SRFBN}& 0.397624 & \textbf{0.627165} & 0.519975 & 0.537452 &  0.359543 & 0.577195 & 0.503159\\
			MTC \cite{MTC}&0.430793 & 0.523213 & \textbf{0.583992} & 0.534206 & 0.404109 & 0.556816 &0.505522 \\
			SPARNet \cite{SPARNet}& 0.444318 & 0.483088 & 0.442757 & 0.543900 & 0.343557 &0.607939 & 0.477593\\
			SISN(Ours) & \textbf{0.475734} & 0.599383 & 0.502253 & \textbf{0.548835} & \textbf{0.490079} & \textbf{0.624352} & \textbf{0.540106} \\
			\hline
		\end{tabular}
	\end{center}
\end{table*}

\subsubsection{Face Recognition}
As a low-level visual task, the ultimate goal of face hallucination is to improve the performance of high-level visual tasks.
Among the high-level vision tasks related to faces, face recognition tasks are widely used in various fields.
Therefore, in this part, we use MTCNN \cite{MTCNN} as the face detector and MobileFacenets \cite{mobilefacenets} as the face recognizer to compare the impact of the reconstructed results of different methods on face recognition performance in real-world surveillance scenarios.

First, we randomly select high-definition frontal face images of six test candidates (identities) captured by SLR (Single Lens Reflex) camera in the SCface dataset as source face feature samples.
Then, we select the LR face images captured by the same surveillance camera corresponding to the six test candidates as the target image of the face recognizer.
To visualize the selected frontal high-definition face image and the face image captured by the surveillance camera, Figure \ref{fig:sample_facerec} shows a sample pair of face images for Candidate 1, where (A) denotes a source face feature sample of Candidate 1, and (B) denotes a target image of the face recognizer.

To take the Candidate 1 for example, the complete face recognition experiment consists of the following steps. First, the high-definition frontal face of Candidate 1 is detected by face detector from (A), and face features are extracted by face recognizer from (A). 
Then the same method is used to perform face detection and feature extraction on (B). 
Finally the face recognizer calculates the similarity based on the two face features to get the normalized similarity (similarity range is 0-1), the higher the value means the face captured by the surveillance camera is more similar to the face of Candidate 1, and further improves the performance of face recognition.
To compare the effect of different face hallucination methods on face recognition performance objectively, we use the LR face image as a baseline and calculate the similarity between the face images reconstructed by different methods (4$\times$ SR) and the high-definition frontal face images.

Table \ref{tab:facerec} shows the face similarity of each LR face images and each HR face images reconstructed by different methods for selected six test candidates after face recognition in real-world surveillance scenarios.
From these quantitative results, we can draw two conclusions: (1) only one face is detected among the six original LR face images; (2) the reconstructed face images after face hallucination can significantly improve the performance of face recognition due to the improved face similarity. 
The above two points prove the challenge of the selected test samples and the importance of face hallucination.
From the data in Table \ref{tab:facerec}, we can further conclude that the proposed method has the highest score on the four test candidates (Candidate 1, 4, 5, and 6) and achieves the best in average similarity, which proves convincingly that the proposed method can maintain the fidelity of facial detail restoration and consistency of facial structure simultaneously.

\section{CONCLUSION}
\label{sec:conclusion}
In this paper, a novel face hallucination method is proposed, which can reconstruct high-quality and high-confidence face images consistently proven by extensive face reconstruction and face recognition experiments.
We design a SISN to maintain the fidelity of facial detail restoration and consistency of facial structure by fusing multi-path features responsible for facial texture details and facial structure information, respectively.
In the future, we believe that the proposed method can be easily applied to other image restoration problems, such as denoising, deblurring and general image SR.

\begin{acks}
This work is supported by the National Natural Science Foundation of China (62072350, 61771353, 62001334), Hubei Technology Innovation Project (2019AAA045), the Central Government Guides Local Science and Technology Development Special Projects (2018ZYYD059), 2020 Hubei Province High-value Intellectual Property Cultivation Project, the Wuhan Enterprise Technology Innovation Project (202001602011971), the Graduate Innovation Fund of Wuhan Institute of Technology (CX2020223).
\end{acks}

\bibliographystyle{ACM-Reference-Format}
\bibliography{refs}


\end{document}